\documentclass{article}

\usepackage[preprint]{neurips_2022}

\usepackage[utf8]{inputenc} 
\usepackage[T1]{fontenc}    
\usepackage{url}            
\usepackage{booktabs}       
\usepackage{amsfonts}       
\usepackage{nicefrac}       
\usepackage{microtype}      
\usepackage{xcolor}         
\usepackage{times}
\usepackage{epsfig}
\usepackage{graphicx}
\usepackage{amsmath}
\usepackage{amssymb}
\usepackage{multirow}
\usepackage{adjustbox}
\usepackage[font=small]{caption}
\usepackage{enumitem}
\usepackage{stmaryrd}
\usepackage{pifont}
\usepackage{xspace}
\usepackage{subcaption}
\usepackage{makecell}
\usepackage[pagebackref=true,breaklinks=true,colorlinks,bookmarks=false]{hyperref}
\usepackage{listings}

\title{Improving the Latent Space of Image Style Transfer}

\author{%
  Yunpeng Bai$^{1}$, Cairong Wang$^{1}$, Chun Yuan$^{1,2}$,  Yanbo Fan$^{3}$, Jue Wang$^{3}$ \\[0.5em]
  $^{1}$Shenzhen International Graduate School, Tsinghua University,\\[0.3em]  $^{2}$Peng Cheng Laboratory,$^{3}$Tencent AI Lab \\[0.3em]
}

\begin{document}

\maketitle

\begin{abstract}

Existing neural style transfer researches have studied to match statistical information between the deep features of content and style images, which were extracted by a pre-trained VGG, and achieved significant improvement in synthesizing artistic images. However, in some cases, the feature statistics from the pre-trained encoder may not be consistent with the visual style we perceived. For example, the style distance between images of different styles is less than that of the same style. In such an inappropriate latent space, the objective function of the existing methods will be optimized in the wrong direction, resulting in bad stylization results. In addition, the lack of content details in the features extracted by the pre-trained encoder also leads to the content leak problem. In order to solve these issues in the latent space used by style transfer, we propose two contrastive training schemes to get a refined encoder that is more suitable for this task. The style contrastive loss pulls the stylized result closer to the same visual style image and pushes it away from the content image. The content contrastive loss enables the encoder to retain more available details. We can directly add our training scheme to some existing style transfer methods and significantly improve their results. Extensive experimental results demonstrate the effectiveness and superiority of our methods. 

\end{abstract}

\section{Introduction}

\label{section1}


Artistic style transfer \cite{DBLP:conf/cvpr/GatysEB16,DBLP:conf/nips/LiFYWLY17,DBLP:conf/cvpr/ParkL19,chen2021artistic,li2019learning,sheng2018avatar} has been a long-term research topic that aims to transfer artistic style from reference image to content image. 
Recent methods \cite{DBLP:conf/cvpr/GatysEB16,DBLP:conf/iccv/HuangB17} use neural networks to match feature statistical information between content and style images. Although these approaches have developed rapidly and achieved significant improvement, there remains a critical problem that has not been discussed: \emph{Is the style forms used by the existing methods, which are based on feature statistics, consistent with the characteristics of visual styles? }

Gatys et al. \cite{DBLP:conf/cvpr/GatysEB16} first proposed the neural style transfer method, which uses image representations derived from a pre-trained Deep Convolutional Neural Network (DCNN) to separate image content from style. In their way, style is defined as the Gram matrix of the deep features, which is related to the fixed parameters of the encoder. Different style definitions are proposed in the later work, but they mostly describe a pre-trained encoder's deep feature statistics (e.g., mean and variance of features \cite{DBLP:conf/iccv/HuangB17}). The encoder pre-trained in different ways get different style representation values for the same image. \emph{So, which encoder is more suitable for style transfer?} Some works \cite{DBLP:conf/wacv/Du20,DBLP:conf/cvpr/WangLV21} have pointed out that even the randomly initialized network can also achieve acceptable style transfer results. This shows that the encoder pre-trained on large collections of images is unnecessary, and it may not even be the most suitable for style transfer.

\begin{figure}[t!]
    \centering
    \includegraphics[width=\linewidth]{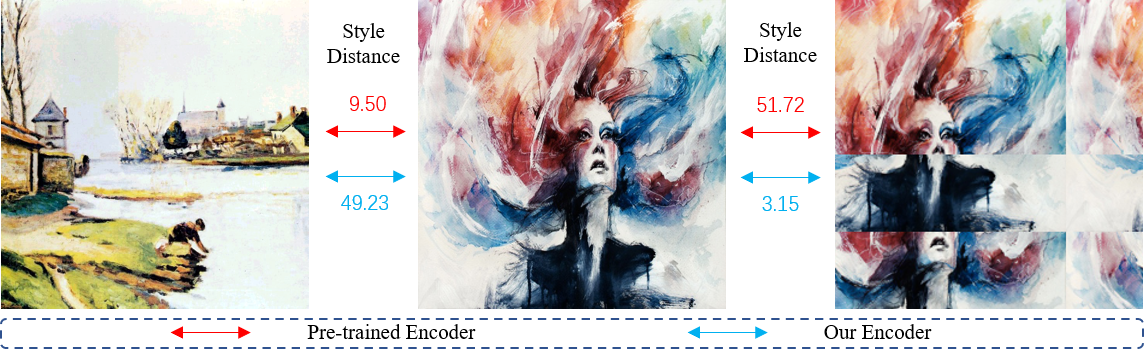}
    \caption{ The two images on the left are randomly selected from the style dataset. We rearrange one of them spatially to get the third image. Then, We use a pre-trained encoder used in previous methods to calculate the style distance for three images. Surprisingly, we find that the distance between the two images on the left was smaller than that of the two images with exactly the same style on the right, which indicates that the objective of the existing methods may not be consistent with the visual style. }
    \label{fig:s_d}   
\end{figure}

On the other hand, we judge the style of images through subjective perception, which may not be consistent with the feature statistics obtained by a pre-trained neural network. As the example shown in Figure \ref{fig:s_d}, we use the method in \cite{DBLP:conf/cvpr/WangLWH020} to calculate the style distance between images with Gram matrix:
\begin{equation}
    \mathcal{D}_\text{style}^{} = \|\mathcal{G}(\mathcal{F}^{}(I_{\text{1}})) - \mathcal{G}(\mathcal{F}^{}(I_{\text{2}}))\|_2.
\label{eqn:style_distance}
\end{equation}
This Gram matrix is obtained by a commonly used pre-trained VGG-19 \cite{DBLP:journals/corr/SimonyanZ14a}. It can be seen that the style distance between different style images is smaller than that of the same style images in the feature space of the pre-trained encoder. Even if we optimize the style loss of existing methods to a smaller value in such an inappropriate latent space, the stylized result may not achieve a visually consistent style. Therefore, we consider optimizing the encoder's parameters while training other modules to make the style representations in its latent space consistent with characteristics of visual styles. Wang et al. \cite{DBLP:conf/cvpr/WangLWH020} try to use the knowledge distillation method to get a new encoder from the pre-trained encoder and achieve better results. However, the new encoder retains the same knowledge as the original encoder in their training process. That is, its latent space has not changed much.

In order to solve this problem, we design a style contrastive training scheme to fine-tune the pre-trained encoder to get a more suitable one, which features statistics that are more style-consistent. By pulling the style positive examples and pushing away the negative examples, the same visual style images will have a more consistent representation in the latent space of the encoder. This training scheme can be directly added to the existing style transfer methods and improve the effect. 

In addition, there is another problem with existing style transfer methods: The stylization results of these CNN-based methods often lose some content information \cite{DBLP:conf/cvpr/AnHSD0L21}. Some works \cite{DBLP:conf/cvpr/ParkL19} try to solve this problem, but they can only train the decoder part to retain more content at most, while ignoring the missing content details in the features extracted by the encoder.  For this, we also propose an identity preserve content contrastive loss to make the encoder retain more local details in fine-tuning. Finally, we conduct experiments on some state-of-the-art style transfer methods and achieved significant improvement.

To summarize, the main contributions of this work are threefold:
\begin{itemize}
\item We propose a style contrastive training scheme to refine the pre-trained encoder used in the existing style transfer method to make its latent space more style-consistent.


\item We propose an identity preserve content contrastive loss to alleviate problem content leak problem caused by the pre-trained encoder.

\item We demonstrate the effectiveness and superiority of our approach by adding our training scheme to some existing methods and achieve significant improvement.
\end{itemize}

\section{Related Work}

\label{section2}
\subsection{Style Transfer} 
Artistic style transfer aims to transfer the style of some artworks to real-world photos, and create a large number of images that have not appeared before. Before the emergence of deep neural network, similar tasks were more like a problem of texture transfer, which mainly tackled by non-parametric sampling \cite{DBLP:conf/iccv/EfrosL99}, non-photorealistic rendering \cite{DBLP:books/daglib/0007390,strothotte2002non} or image analogy \cite{DBLP:conf/siggraph/HertzmannJOCS01}. With the help of DCNN, Gatys et al. \cite{DBLP:conf/cvpr/GatysEB16} first propose the neural style transfer, which uses deep features from the pre-trained network to represent style and content. In their way, style is defined as the Gram matrix of deep features, and stylization is achieved by matching the second-order statistics between the result and style image. In the later work, different style definitions \cite{DBLP:conf/iccv/HuangB17} are proposed, while they mostly describe the statistics of deep features from a pre-trained encoder. However, these forms may not be consistent with the image's style, which is caused by the inappropriate encoder. Recently, some works have explored the influence of the encoder on style transfer, such as the encoder parameters \cite{DBLP:conf/wacv/Du20} and encoder architecture \cite{DBLP:conf/cvpr/WangLV21}. Wang et al. \cite{DBLP:conf/cvpr/WangLWH020} try to use the knowledge distillation method to get a new encoder from the pre-trained encoder and achieve better results. However, the new encoder retains the same knowledge as the original encoder in their training process. That is, its latent space has not changed much. In contrast, we try to improve the latent space of the encoder to make it more style-consistent and apply the training scheme to the existing methods.

\subsection{Contrastive Learning}

Contrastive learning first appeared in the field of unsupervised representation learning and has shown great promise\cite{DBLP:conf/cvpr/He0WXG20}. These methods are based on the theory of maximizing mutual information. The basic idea is to build an embedding space where associated signals are pulled together while other samples in the dataset are pushed away. 
Signals may vary depending on specific tasks. A new form of contrastive loss called InfoNCE \cite{DBLP:journals/corr/abs-1807-03748}, which measures the similarity by dot production, is proposed as a representative loss function to maximize a lower bound of the mutual information.
Later, the effectiveness of contrastive learning was gradually verified on various tasks, such as semantic segmentation \cite{DBLP:conf/iccv/WangZYDKG21}, object detection  \cite{xie2021detco} and classification \cite{wang2021contrastive}.

In the field of conditional image synthesis, contrastive learning has also received extensive attention \cite{park2020contrastive,yu2021dual}. More recently, Liu et al. \cite{liu2021divco} introduced a latent-augmented contrastive loss to achieve diverse image synthesis. Wu et al. \cite{wu2021contrastive} improved the image dehazing result by pulling the restored image closer to the clear image and pushing it far away from the hazy image. Chen et al. \cite{chen2021artistic} first adapt contrastive learning to the artistic style transfer to learn so-called stylization-to-stylization relations. However, suppose the style transfer process is carried out in an unsuitable latent space. In that case, the stylization-to-stylization relations are meaningless, because they may all not be visually consistent with the style image. In contrast, we use the contrastive learning method to make the same visual style images have consistent representations in the latent space.

\subsection{Knowledge Distillation}
Knowledge distillation (KD) \cite{ba2014deep, hinton2015distilling,yu2019learning} is a model compression method, in which a student network is trained by learning the knowledge from a teacher network. The knowledge is expressed in the form of softened probability \cite{yu2019learning,peng2019correlation}, which can reflect the inherent class similarity structure known as dark knowledge. The distillation objective encourages the output probability distribution over predictions from the student and teacher networks to be similar. With the help of additional information on top of the one-hot labels, the performance of student network can be boosted. This dark knowledge is mainly related to labels, so they are rarely used in low-level vision tasks (e.g., neural style transfer). Wang et al. \cite{DBLP:conf/cvpr/WangLWH020} developed a collaborative knowledge distillation method to learn a much smaller model from pre-trained redundant VGG-19 for ultra-resolution style transfer. In our method, the pre-trained encoder is regarded as a regularizer to guarantee that the features extracted by the new encoder are near a suitable value.

\begin{figure}[t!]
    \centering
    \includegraphics[width=\linewidth]{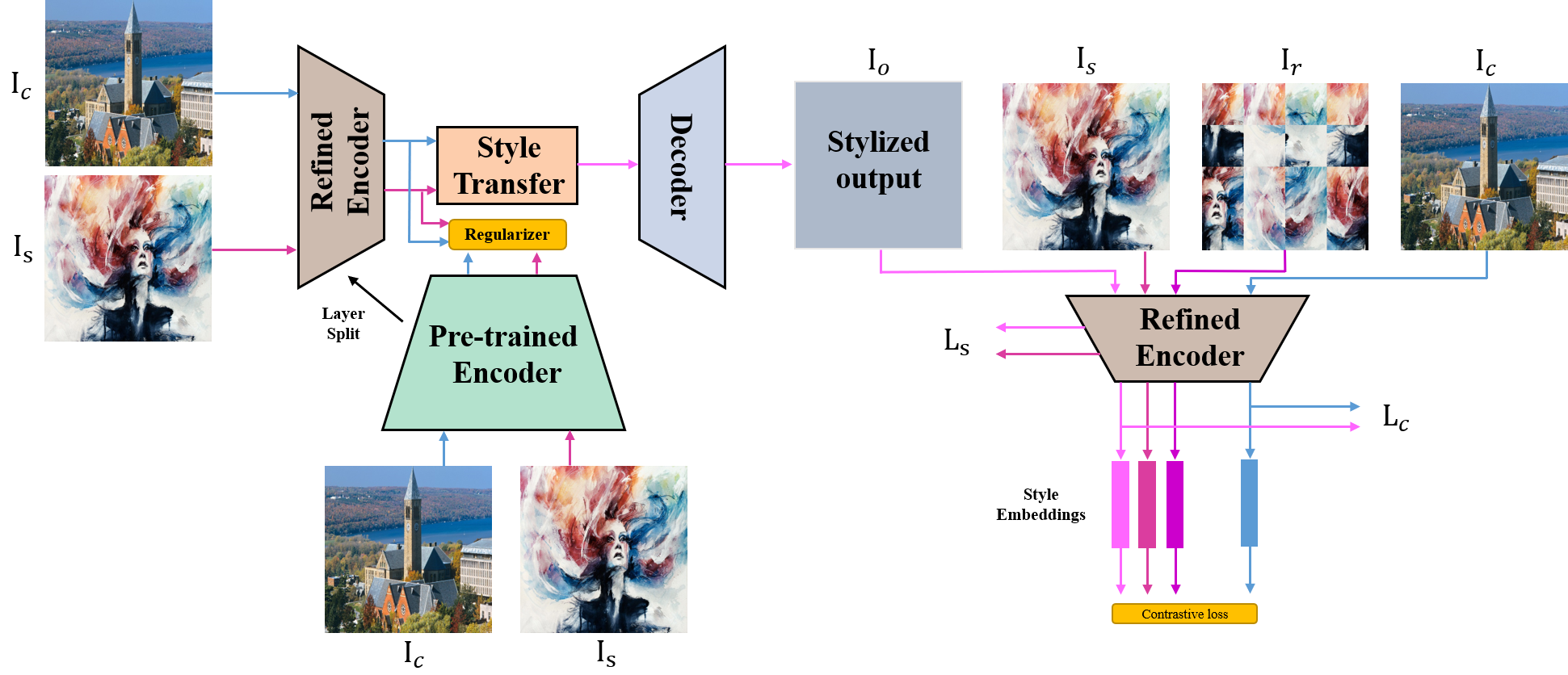}
    \caption{ Our style contrastive training scheme. For some learning-based style transfer methods that use encoder-decoder architecture, we optimize their encoders in the training process simultaneously. The original pre-trained encoder is used as a feature regularizer. The stylized result is pulled closer to the same visual style image and pushed far away from the content image in the refined encoder’s latent space. In the figure, We use similar colors to represent the embeddings of the same style. The style loss $L_s$ and the content loss $L_c$ remain the same as the original method. }
    \label{fig:arch1}   
\end{figure}

\section{Proposed Method}
\label{section3}
\vspace{-0.1in}
Arbitrary style transfer methods \cite{DBLP:conf/iccv/HuangB17,DBLP:conf/cvpr/ParkL19,chen2021artistic} typically adopt an encoder-decoder architecture, which transfers the style in the encoder feature space, and then converts them back to the stylized results through the decoder. The encoder often adopts a pre-trained VGG-19 \cite{DBLP:journals/corr/SimonyanZ14a} to extract expressive informative representations. In the training process, the parameters of the encoder are fixed to provide supervision signals. However, due to the style inconsistency and lack of content details in the latent space of the pre-trained encoder, the quality of stylization results is affected.
We try to refine this encoder to make it more suitable for the style transfer task.

\subsection{Encoder Fine Tuning}
\vspace{-0.1in}
We refine the encoder by optimizing its parameters while training other modules. If we directly set the encoder's parameters to be optimized, it is equivalent to optimizing the output result and supervision signals simultaneously and will get degraded results. To avoid this problem, we use the pre-trained encoder as a feature regularizer, which prevents the new features from being too far away from a suitable scale. The new encoder uses the same architecture and is initialized with the parameters of the pre-trained one, which is easier to converge than random initialization. Further, we find that some layers in the original architecture are unnecessary. Therefore, we propose a layer split scheme, which only retains the necessary layers in the original encoder. On this basis, we further add two contrastive training schemes to improve the encoder to make it more suitable for the style transfer task.


\begin{figure}[t!]
    \centering
    \includegraphics[scale=0.45]{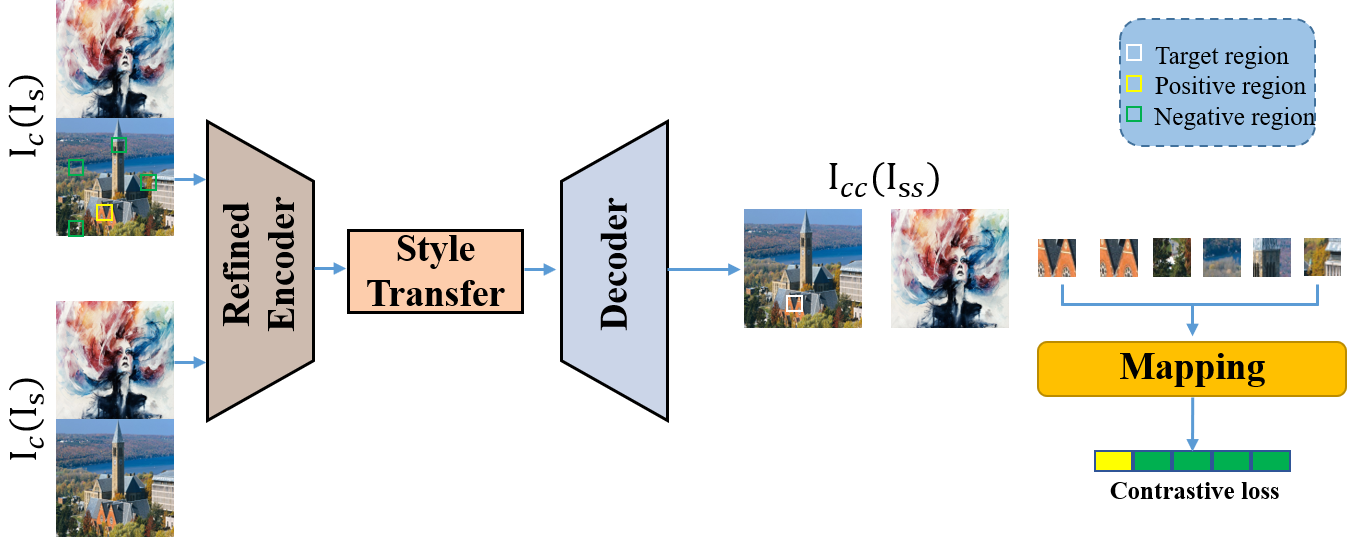}
    \caption{ Our identity preserve content contrastive loss. $I_c$ ($I_s$) is the input content (style) image and $I_{cc}$($I_{ss}$) is the output image synthesized from this image pair (content or style). We sample some blocks of the restored and original images and compare them in a latent space. The features of the same position are pulled together, and different positions are pushed away. In this way, the encoder can retain more available details when extracting features. }
    \label{fig:arch2}   
\end{figure}

\subsection{Style Contrastive Loss}
\vspace{-0.1in}
Contrastive learning is widely used in self-supervised representation learning, which is orthogonal to the training method of style transfer. We use the contrastive learning method to make the images with the same visual style have similar representations. As shown in Figure \ref{fig:arch1}, in our scheme, the stylized result is pulled closer to the same visual style “positive” examples and pushed far away from the “negative” examples in the refined encoder’s latent space.  If only the original style image served as the positive example to be pulled, some content in the result would be lost. This is because the learned style embeddings are not well decoupled from the content. The content of the result is also pulled closer to the style image. To solve this problem, we design more style positive examples, and then also pull the result with these examples in the latent space. These examples are obtained by spatial rearrangement of the style image. Specifically, the style image is divided into $n * n $ blocks, which are then randomly disrupted and recombined to obtain the images $\{I_{r}^{0}, I_{r}^{1}, \dots ,I_{r}^{N} \}$. We assume that these recombined images share the same style but different content as the original image. The features of these positive images and the stylized result from the refined encoder will be mapped into embeddings through a mapping network. We take the content images (including the original content image) loaded into the batch as negative examples, and they are also mapped to embeddings through the same network. Then we can formulate our style contrastive loss as follows:

\begin{equation}
\mathcal{L}_{s-c}= \sum_i -\log \left(\frac{\exp \left(f_{s}\left(E(I_o)\right)^{T} f_{s}\left(E(I_{r}^{i})\right) / \tau\right)}{\exp \left(f_{s}\left(E(I_o)\right)^{T} f_{s}\left(E(I_{r}^{i})\right) / \tau\right)+\sum_j \exp \left(f_{s}\left(E(I_o)\right)^{T} f_{s}\left(E(I_{c}^{j})\right) / \tau\right)}\right),
\end{equation}
where $I_o, I_{r} $ and $I_{c}$ represent the output result, reshuffled style image and content image, respectively. $E$ is our refined encoder. $f_{s}$ is the style mapping network which used to obtain the style embedding consistent with the visual style of the image. $\tau$ is a temperature hyper-parameter to control push and pull force. 

\subsection{Identity Preserve Content Contrastive Loss}
\vspace{-0.1in}
The content leak in the style transfer results has been noticed \cite{DBLP:conf/cvpr/ParkL19,DBLP:conf/cvpr/AnHSD0L21}, which remains a problem to be solved. Some existing methods try to use an identity loss \cite{chen2021artistic} to preserve the content structure, but they can only train the decoder to retain the content at most, while ignoring the content leak in the encoder part. In addition, their identity losses only introduce a reconstruction loss or perceptual loss between the output image synthesized from two same content (or style) images with the original image to keep the structure. The effect is not apparent because both kinds of loss do not emphasize the local region where the content is more likely to lose.


In order to alleviate this problem, we design the identity loss as a local-wise contrastive loss similar to \cite{park2020contrastive}, which enables the encoder to retain more details when extracting features. Same as SANet, we use content (or style) features to stylize itself and obtain a restored content image. Then, as shown in Figure \ref{fig:arch2}, we randomly select several blocks from the same positions of the restored image and the content image. Then, these image blocks will be mapped into latent codes that encode local structures through the same mapping network. A contrastive loss is introduced to make the latent from the same position pulled together and pushed away from other positions. Such a local-wise identity preserve content contrastive loss is expressed as:

\begin{equation}
\mathcal{L}_{c-c}= \sum_i -\log \left(\frac{\exp \left(f_{c}\left(p_{cc}^i\right)^{T} f_{c}\left(p_{c}^{i}\right) / \tau\right)}{\exp \left(f_{c}\left(p_{cc}^i\right)^{T} f_{c}\left(p_{c}^{i}\right) / \tau\right)+\sum_{j\neq i} \exp \left(f_{c}\left(p_{cc}^{i}\right)^{T} f_{c}\left(p_{c}^{j}\right) / \tau\right)}\right),
\end{equation}

where $p_{cc}$ denotes the random blocks of the output image synthesized from two same content images. Similarly, the $p_{c}$ denotes the blocks of original content image. $f_{c} = h_{c}\left(\phi_{{i}}(\cdot)\right)$, where $\phi_{i}$ denotes a $\verb+ReLU_X_1+$ layer in our refined encoder and $h_{c}$ is a content mapping network.

\subsection{Objective Function}
\vspace{-0.1in}
We can directly add our training scheme to some existing encoder-decoder-based style transfer methods without changing the original architectures, and significantly improve their effect. The pre-trained encoder used in the original method can be fine-tuned by optimizing the following function:


\begin{equation}
\mathcal{L}_{total} = \underbrace {\mathcal{L}_{ast}}_\text{original loss} +\underbrace{\lambda_d \sum_{i=1}^k  \|\mathcal{F}^{(i)}_n - \mathcal{F}^{(i)}_o\|_2}_\text{feature regularizer} +\underbrace{\lambda_{s-c}\mathcal{L}_{c-c } +  \lambda_{s-c}\mathcal{L}_{s-c }}_\text{feature improvement}
\label{eq:objective}    
\end{equation}
where ${L}_{ast}$ is the loss function of the original method, $\mathcal{F}_n$ is the feature extracted by the refined encoder, and $\mathcal{F}_o$ is that of the pre-trained one. $\lambda_{d}$, $\lambda_{s-s}$ and $\lambda_{s-c}$ are the corresponding loss weights, respectively.

\begin{figure}[t!]
    \centering
    \includegraphics[width=\linewidth]{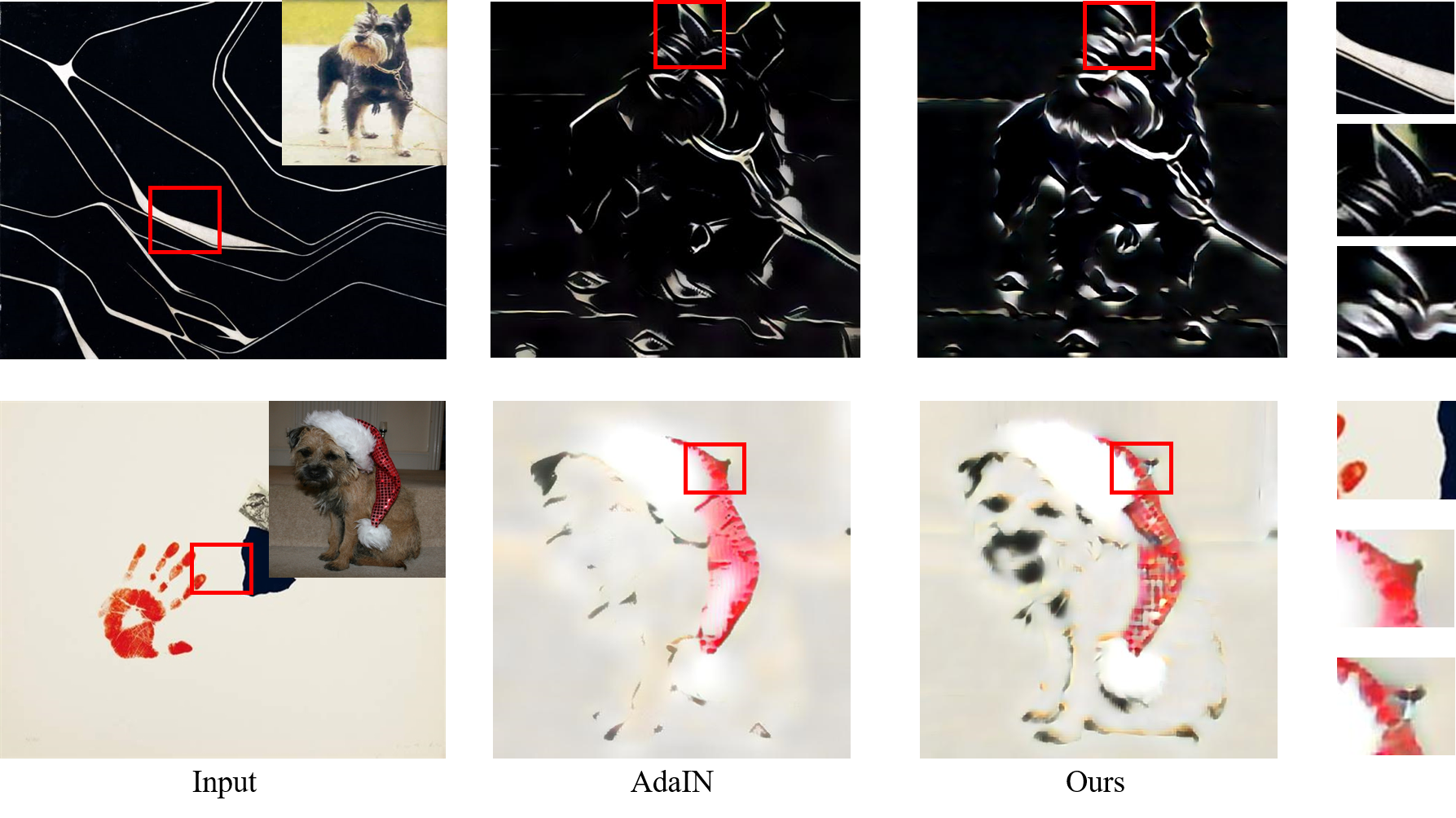}
    \caption{ Experiment on AdaIN }
    \label{fig:adain}   
    \vspace{-2mm}
\end{figure}

\section{Experimental Results}
\label{section4}
\vspace{-0.1in}
In this section, we first introduce some implementation details, then add our training scheme to several existing methods and show the improvement on their methods. We also make further comparisons between our method and several baseline models. Finally, we explored the effect of our contrastive training scheme through ablation studies, especially the number of positive and negative examples.

\begin{table}
    \centering
    \caption{ The user study scores for different methods. The higher the better.}
    \begin{tabular}{l|c|c|c|c}
    \toprule
       Stylization scheme                         & AdaIN             & Our AdaIN & SANet & Our SANet \\
    \midrule
        Preference Score               & 0.153              & 0.228         & 0.242 & \textbf{0.377} \\

    \bottomrule
    \end{tabular}
    \label{tab:user_study}
\end{table}

\subsection{Implementation Details}
\vspace{-0.1in}
The datasets for our experiments are the commonly used MS-COCO \cite{lin2014microsoft} (for the content images) and WikiArt \cite{lin2014microsoft} (for the style images). Both datasets contain roughly $80,000$ training images. The optimizer (usually Adam \cite{kingma2014adam}) and the learning rate are the same as the corresponding methods. In style contrastive loss, the number of style positive examples is set to 8, and the number of negative examples is the same as batch size. The style mapping network $f_s$ consists of a convolution layer and several subsequent MLP (multi-layer perceptron) layers, of which the last MLP layer has 128 units. The content mapping network $h_c$ is a two-layer MLP. The number of units in the first layer is the same as the corresponding feature channel, and the second layer has $256$ units. The hyper-parameter $\tau$ is set to $0.07$ in both contrastive losses. The image is also processed as the previous methods: the smaller dimension of the two images is rescaled to $512$ while retaining the aspect ratio and then randomly cropped to the size of $256*256$ pixels. Our new refined encoder is still VGG-style. For our layer split scheme, we only retain \verb+Conv_X_1+ layers and the corresponding \verb+pooling+ and \verb+ReLU+ layers of the original VGG-19. 

\subsection{Experiments on Existing Methods}
\vspace{-0.1in}
\textbf{Experiment on AdaIN.} The AdaIN \cite{DBLP:conf/iccv/HuangB17} method presents an efficient solution for universal style transfer. It receives a content input $x$ and a style input $y$, and simply aligns the channel wise mean and variance of content feature maps to those of style feature maps as:
\begin{equation}
\operatorname{AdaIN}(x, y)=\sigma(y)\left(\frac{x-\mu(x)}{\sigma(x)}\right)+\mu(y)
\end{equation}
The features used here are the \verb+ReLU_4_1+ layer features extracted by a pre-trained VGG-19 encoder.
After the style swap operation in the feature space, the output of AdaIN can be inverted to the image space with a feed-forward decoder to get the final output. On their basis, our new encoder only retains several convolution layers from the input layer to \verb+ReLU_4_1+. In the training process, the regular term is only added between the activation values of \verb+ReLU_4_1+ layers of the two networks. In content contrastive loss, the $\phi$ we used are \verb+ReLU_1_1+, \verb+ReLU_2_1+, \verb+ReLU_3_1+ and \verb+ReLU_4_1+. The comparison results are shown in Figure \ref{fig:adain}. Because AdaIN uses an inappropriate encoder, the final synthesized result retains the original color or the color that has not appeared, and a large amount of content information is lost. In contrast, our results retain more content and are more style-consistent.

\begin{figure}[t!]
    \centering
    \includegraphics[width=\linewidth]{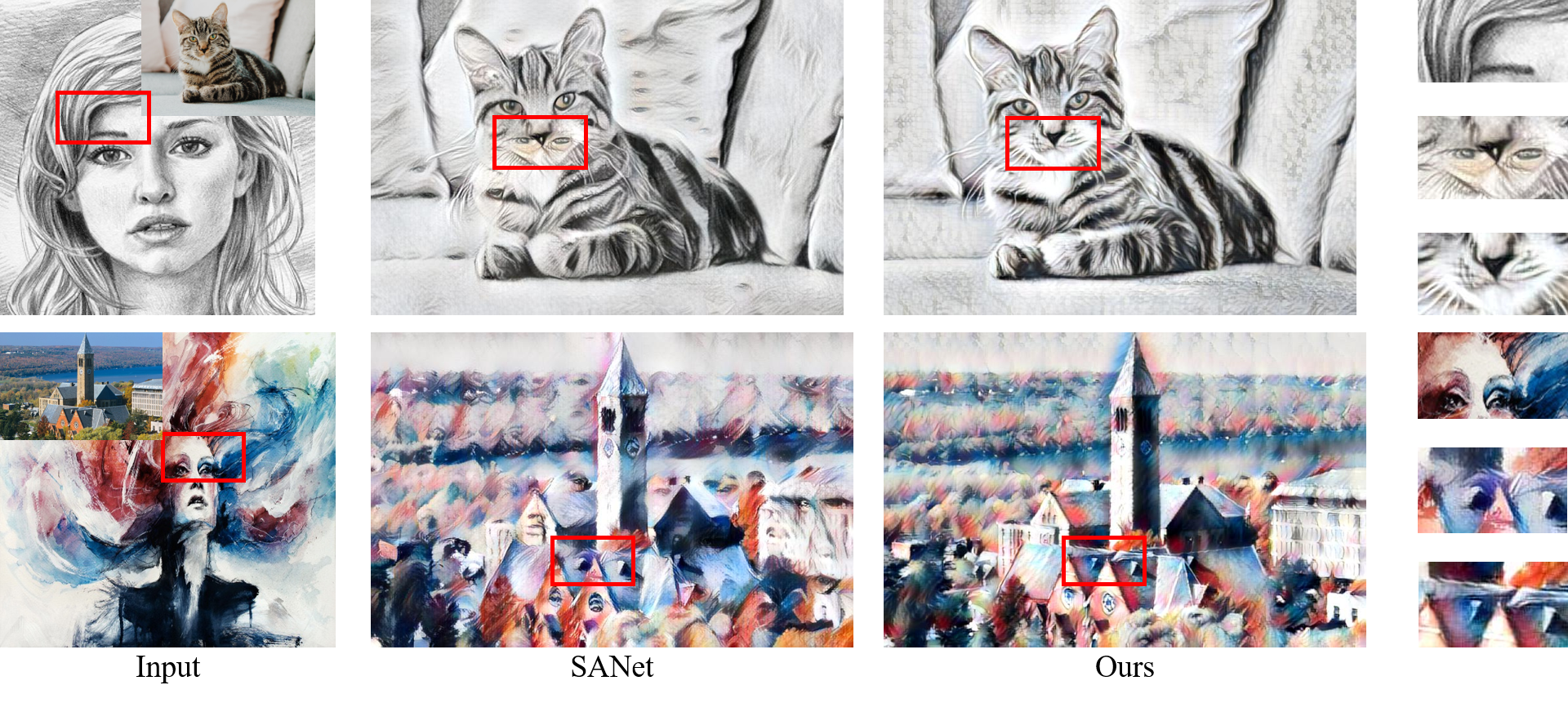}
    \caption{ Experiment on SANet.}
    \label{fig:sanet}   
\end{figure}

\begin{figure}[t!]
    \centering
    \includegraphics[width=\linewidth]{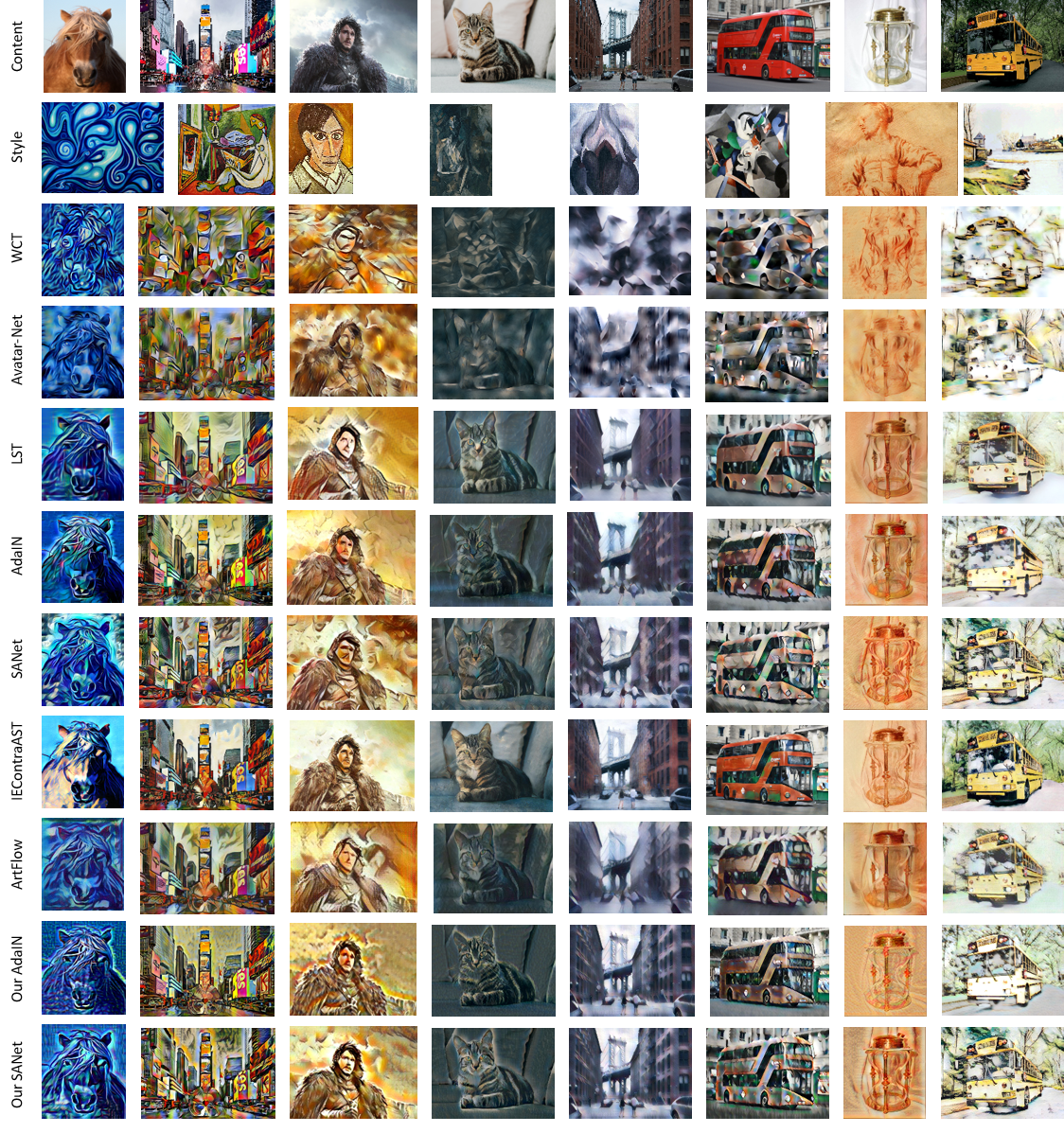}
    \caption{ Comparisons with other style transfer methods. }
    \label{fig:comparison}   
\end{figure}


\textbf{Experiment on SANet.} SANet \cite{DBLP:conf/cvpr/ParkL19} also follows the encoder-decoder architecture, where the transfer part consists of two style-attention networks. After encoding the content and style images by the pre-trained VGG encoder, the SANet maps features from \verb+ReLU_4_1+ and \verb+ReLU_5_1+ features. In our training scheme, the encoder adds the layers from \verb+Conv_5_1+ to \verb+ReLU_5_1+ compared with the above AdaIN scheme, and the regular term is added on the \verb+ReLU_4_1+ and \verb+ReLU_5_1+ layers. The content contrastive loss is the same as AdaIN above. As shown in Figure \ref{fig:sanet}, SANet is easy to migrate the patch with semantic information in the style image, resulting in strange results. In contrast, our method can ensure the correct content structure.

\textbf{User Study.} For a more objective comparison, a user study was also conducted.
We generate 100 stylized images using each model. These images were presented to $50$ participants in random order. Participants were asked to choose their favorite image for each content-style pair. The user study results are shown in Table ~\ref{tab:user_study}. Our method's stylized images are more preferred than those of the original methods.

\textbf{Comparisons with Other Methods.}
We also compare our results with existing methods, including AdaIN \cite{DBLP:conf/iccv/HuangB17}, WCT \cite{DBLP:conf/nips/LiFYWLY17}, Avatar-Net \cite{sheng2018avatar}, LST \cite{li2019learning}, SANet\cite{DBLP:conf/cvpr/ParkL19}, Artflow\cite{DBLP:conf/cvpr/AnHSD0L21} and IEContraAST\cite{chen2021artistic}. Results of these methods are obtained by using the public codes and default configurations and are shown in Figure \ref{fig:comparison}. The learning-free methods \cite{DBLP:conf/nips/LiFYWLY17,sheng2018avatar} cannot separate style and content well, so they often fail to preserve the content structure and get distorted stylized images. Because IEcontroAST\cite{chen2021artistic} relies on external learning, the stylized result will deviate from the style of the style image or even no style transferred at all. The flow model used in ArtFlow  \cite{chen2021artistic} can not guarantee that the content structure is not distorted. Our method makes the stylized results achieve a better trade-off between style and content through two training schemes, which is to keep the content details of the original image while keeping the style consistent with the style picture as much as possible.






\subsection{Ablation Studies}
\begin{figure}[t!]
    \centering
    \includegraphics[width=\linewidth]{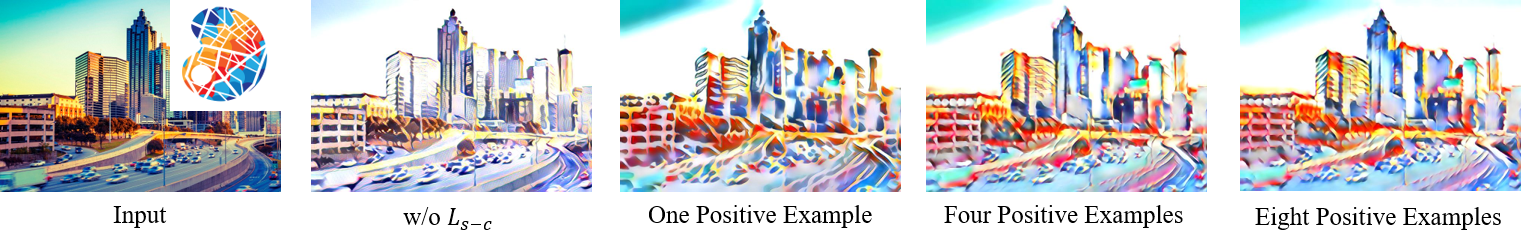}
    \caption{Abalation study on number of style positive examples used in style contrastive learning.  }
    \label{fig:ablation}   
    \vspace{-0.1in}
\end{figure}

\begin{figure}[t!]
    \centering
    \includegraphics[width=\linewidth]{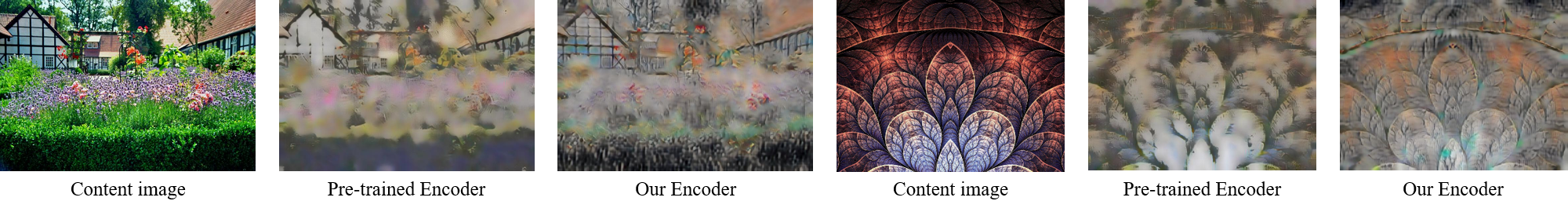}
    \caption{ Content visualization. }
    \label{fig:content}   
    \vspace{-0.1in}
\end{figure}

\vspace{-0.1in}
We first study the effect of the number of examples on the results in style contrastive learning. As shown in Figure \ref{fig:ablation}, when there is only one positive example, the style embedding is not well decoupled from the content, resulting in distortion in some areas. With the increasing number of positive examples, this problem will be alleviated and basically solved when set to 8. When the $L{s-c}$ is removed, the style of the result is inconsistent with the style image. Although more negative examples usually lead to better performance in contrastive learning, the most crucial negative example in the style transfer task is the original content image (to distinguish from the original image style). Adding more negative examples on the basis of the original content image does not make much sense. Further, in order to prove that our training method can well preserve the content in the features extracted by the encoder, we made a visualization for the content feature. As shown in Figure \ref{fig:content}, it can be seen that some very fine details are not lost in our encoder. Moreover, our training scheme can well alleviate the inconsistent visual style problem in the pre-trained encoder. In addition to the example shown in Figure \ref{fig:s_d}, see our supplementary file for more results.









\section{Discussion}
\label{section5}
\vspace{-0.1in}
\textbf{Limitations.} 
Because our training scheme needs to be added to the existing learning-based style transfer methods(such as AdaIN, SANet, etc.), and the decoder in the architecture is necessary to convert the transfer results back to the image space for content contrastive loss, our method can not be applied to some optimization-based methods. Another limitation is that some style positive examples need to be loaded in the process of contrastive training, which will reduce the efficiency of training. In our future work, we will explore a pre-training scheme to get an encoder that can be directly applicable to various style transfer tasks.

\textbf{Conclusions.} 
In this work, we study the irrationality of the pre-trained VGG encoder used in existing style transfer methods, which shows that the style distance of some images with different styles in the feature space of the pre-trained encoder is less than that of images with the same style. 
This means that the encoder used in the previous method to provide supervision signals can not give a visually consistent style representation, resulting in failed style transfer result in some cases. Two contrastive training schemes are proposed to improve the encoder's latent space of the existing work, make it more style-consistent and retain more available details. Extensive experiments show the effectiveness and superiority of our method. Furthermore, this study shows that the pre-trained encoder can be replaced with a more appropriate choice in the later style transfer research.

{\small
\bibliography{egbib}
}

\appendix




\end{document}